\documentclass[runningheads]{llncs}
\usepackage{graphicx}
\usepackage{amsmath,amssymb} 
\usepackage{color}
\usepackage{array,multirow,booktabs}
\usepackage{subcaption}
\usepackage{caption}
\graphicspath{{./pics/}}
\usepackage[outdir=./pics/]{epstopdf}
\usepackage{pifont}
\usepackage{hyperref}
\hypersetup{
    colorlinks=true,
    urlcolor=magenta,
}
\newcommand{\cmark}{\ding{51}}%
\newcommand{\xmark}{\ding{55}}%

\newcommand{\dataset}{YouTube-VOS}
\DeclareMathOperator*{\argmax}{arg\,max}

\usepackage{xspace}
\makeatletter
\DeclareRobustCommand\onedot{\futurelet\@let@token\@onedot}
\def\@onedot{\ifx\@let@token.\else.\null\fi\xspace}

\def\eg{\emph{e.g}\onedot} 
\def\ie{\emph{i.e}\onedot} 
 
\def\etc{\emph{etc}\onedot} 
 
\def\etal{\emph{et al}\onedot}
\makeatother

\begin{document}


\title{~\dataset: Sequence-to-Sequence Video Object Segmentation} 

\titlerunning{~\dataset}

\author{Ning Xu\inst{1} \and
Linjie Yang\inst{2} \and
Yuchen Fan\inst{3}\and
Jianchao Yang\inst{2}\and
Dingcheng Yue\inst{3} \and
Yuchen Liang\inst{3} \and
Brian Price\inst{1} \and
Scott Cohen\inst{1} \and
Thomas Huang\inst{3}}

\authorrunning{N. Xu~\etal}



\institute{Adobe Research, USA \\
\email{{\{nxu,bprice,scohen\}}}@adobe.com \and
Snapchat Research, USA \\
\email{\{linjie.yang,jianchao.yang\}@snap.com}\\ \and
University of Illinois at Urbana-Champaign, USA\\
\email{\{yuchenf4,dyue2,yliang35,t-huang1\}@illinois.edu}}

\maketitle

\begin{abstract}
Learning long-term spatial-temporal features are critical for many video analysis tasks. However, existing video segmentation methods predominantly rely on static image segmentation techniques, and methods capturing temporal dependency for segmentation have to depend on pretrained optical flow models, leading to suboptimal solutions for the problem. End-to-end sequential learning to explore spatial-temporal features for video segmentation is largely limited by the scale of available video segmentation datasets, i.e., even the largest video segmentation dataset only contains 90 short video clips. To solve this problem, we build a new large-scale video object segmentation dataset called YouTube Video Object Segmentation dataset (\dataset). Our dataset contains 3,252 YouTube video clips and 78 categories including common objects and human activities\footnote{This is the statistics when we submit this paper, see updated statistics on our website.}. This is by far the largest video object segmentation dataset to our knowledge and we have released it at \href{https://youtube-vos.org}{https://youtube-vos.org}. 
Based on this dataset, we propose a novel sequence-to-sequence network to fully exploit long-term spatial-temporal information in videos for segmentation. We demonstrate that our method is able to achieve the best results on our \dataset~test set and comparable results on DAVIS 2016 compared to the current state-of-the-art methods. Experiments show that the large scale dataset is indeed a key factor to the success of our model.

\keywords{Video Object Segmentation, Large-scale Dataset, Spatial-Temporal Information.}
\end{abstract}

\section{Introduction}

Learning effective spatial-temporal features has been demonstrated to be very important for many video analysis tasks. For example, Donahue \etal~\cite{donahue2015rcn} propose long-term recurrent convolution network for activity recognition and video captioning. Srivastava \etal\cite{srivastava2015unsupervisedlstm} propose unsupervised learning of video representation with a LSTM autoencoder. Tran \etal~\cite{tran2015c3d} develop a 3D convolutional network to extract spatial and temporal information jointly from a video. Other works include learning spatial-temporal information for precipitation prediction~\cite{xingjian2015convolutional}, physical interaction~\cite{finn2016videopred}, and autonomous driving~\cite{xu2016driving}. 

Video segmentation plays an important role in video understanding, which fosters many applications, such as accurate object segmentation and tracking, interactive video editing and augmented reality. Video object segmentation, which targets at segmenting a particular object instance throughout the entire video sequence given only the object mask on the first frame, has attracted much attention from the vision community recently~\cite{Caelles2017osvos,Perazzi2017masktrack,Yang2018osmn,Cheng2017segflow,Jain_2017_CVPR,Jampani2017vpn,Tokmakov2017memory,Hu2017Maskrnn,voigtlaender2017online}. However, existing state-of-the-art video object segmentation approaches primarily rely on single image segmentation frameworks~\cite{Caelles2017osvos,Perazzi2017masktrack,Yang2018osmn,voigtlaender2017online}. For example, Caelles \etal~\cite{Caelles2017osvos} propose to train an object segmentation network on static images and then fine-tune the model on the first frame of a test video over hundreds of iterations, so that it remembers the object appearance. The fine-tuned model is then applied to all following individual frames to segment the object without using any temporal information. Even though simple, such an online learning or one-shot learning scheme achieves top performance on video object segmentation benchmarks~\cite{Perazzi2016davis,Jain2014youtubeobjects}. Although some recent approaches~\cite{Jain_2017_CVPR,Cheng2017segflow,Tokmakov2017memory} have been proposed to leverage temporal consistency, they depend on models pretrained on other tasks such as optical flow~\cite{Ilg2017flownet,Revaud2015epicflow} or motion segmentation~\cite{Tokmakov2017mpnet}, to extract temporal information. These pretrained models are learned from separate tasks, and therefore are suboptimal for the video segmentation problem.

Learning long-term spatial-temporal features directly for video object segmentation task is, however, largely limited by the scale of existing video object segmentation datasets. For example, the popular benchmark dataset DAVIS~\cite{Pont-Tuset2017davis} has only 90 short video clips, which is barely sufficient to learn an end-to-end model from scratch like other video analysis tasks. Even if we combine all the videos from available datasets~\cite{Jain2014youtubeobjects,jumpcut,Fli2013segtrack,brox2010BMS,ochs2014FBMS,galasso2013VSB100}, its scale is still far smaller than other video analysis datasets such as YouTube-8M~\cite{abu2016youtube} and ActivityNet~\cite{heilbron2015activitynet}. To solve this problem, we present the first large-scale video object segmentation dataset called~\dataset~(YouTube Video Object Segmentation dataset) in this work. Our dataset contains 3,252 YouTube video clips featuring 78 categories covering common animals, vehicles, accessories and human activities. Each video clip is about 3$\sim$6 seconds long and often contains multiple objects, which are manually segmented by professional annotators. Compared to existing datasets, our dataset contains a lot more videos, object categories, object instances and annotations, and a much longer duration of total annotated videos. Table~\ref{tab:dataset-cmp} provides quantitative scale comparisons of our new dataset against existing datasets. We retrain existing algorithms on~\dataset~and benchmark their performance on our test set which contains 322 videos. In addition, our test set contains 10 categories unseen in the training set and are used to evaluate the generalization ability of existing approaches. 

Based on Youtube-VOS, we propose a new sequence-to-sequence learning algorithm to explore spatial-temporal modeling for video object segmentation. We utilize a convolutional LSTM~\cite{xingjian2015convolutional} to learn long-term spatial-temporal information for segmentation. At each time step, the convolutional LSTM accepts last hidden states and an encoded image frame, it then outputs encoded spatial-temporal features which are decoded into a segmentation mask. Our algorithm is different from existing approaches in that it fully exploits the long-term spatial-temporal information in an end-to-end manner and does not depend on existing optical flow or motion segmentation models. We evaluate our algorithm on both~\dataset~and DAVIS 2016 and it achieves better or comparable results compared to the current state of the arts.

The rest of our paper is organized as follows. In Section~\ref{sec:related} we briefly introduce the related works. In Section~\ref{sec:dataset} and~\ref{sec:algorithm} we describe our \dataset~dataset and the proposed algorithm in detail. Experimental results are presented in Section~\ref{sec:exp}. Finally we conclude the paper in Section~\ref{sec:conclusion}. 

\begingroup
\setlength{\tabcolsep}{1.4pt}
\begin{table}[t]
\footnotesize
\centering
\caption{ Scale comparison between~\dataset~and existing datasets. ``Annotations'' denotes the total number of object annotations. ``Duration'' denotes the total duration (in minutes) of the annotated videos. }
\label{tab:dataset-cmp}
\begin{tabular}{|l|c|c|c|c|c|c|c|}
\hline
Scale & \begin{tabular}[x]{@{}c@{}}{JC}\\\cite{jumpcut}\end{tabular} & \begin{tabular}[x]{@{}c@{}}{ST}\\\cite{Fli2013segtrack}\end{tabular}   & \begin{tabular}[x]{@{}c@{}}{YTO}\\\cite{Jain2014youtubeobjects}\end{tabular} &  \begin{tabular}[x]{@{}c@{}}{FBMS}\\\cite{ochs2014FBMS}\end{tabular} & \multicolumn{2}{c|}{\begin{tabular}[x]{@{}c@{}}\multicolumn{1}{c}{DAVIS}\\\cite{Perazzi2016davis}~~\cite{Pont-Tuset2017davis}\end{tabular} }  & \begin{tabular}[x]{@{}c@{}}{\textbf{~\dataset}}\\(\textbf{Ours})\end{tabular}\\
\hline
Videos & 22 & 14 & 96 & 59 & 50 & 90 & \textbf{3,252} \\
\hline
Categories & 14 & 11 & 10 & 16 & - & - & \textbf{78} \\ 
\hline
Objects & 22 & 24 & 96 & 139 & 50 & 205 & \textbf{6,048} \\ 
\hline
Annotations & 6,331 & 1,475 & 1,692 & 1,465 & 3,440 & 13,543 & \textbf{133,886}  \\
\hline
Duration & 3.52 & 0.59 & 9.01 & 7.70 & 2.88 & 5.17 & \textbf{217.21}  \\
\hline
\end{tabular}
\end{table}
\setlength{\tabcolsep}{1.4pt}
\endgroup

\section{Related work} \label{sec:related}

In the past decades, several datasets~\cite{Jain2014youtubeobjects,jumpcut,Fli2013segtrack,brox2010BMS,ochs2014FBMS,galasso2013VSB100} have been created for video object segmentation. All of them are in small scales which usually contain only dozens of videos. In addition, their video content is relatively simple (\eg no heavy occlusion, camera motion or illumination change) and sometimes the video resolution is low. Recently, a new dataset called DAVIS~\cite{Perazzi2016davis,Pont-Tuset2017davis} was published and has become the benchmark dataset in this area. Its 2016 version contains 50 videos with a single foreground object per video while the 2017 version has 90 videos with multiple objects per video. In comparison to previous datasets~\cite{Jain2014youtubeobjects,jumpcut,Fli2013segtrack,brox2010BMS,ochs2014FBMS,galasso2013VSB100}, DAVIS has both higher-quality of video resolutions and annotations. In addition, their video content is more complicated with multi-object interactions, camera motion, and occlusions.

Early methods~\cite{Jain2014youtubeobjects,Nagaraja2015video,Faktor2014voting,papazoglou2013fast,brox2010object} for video object segmentation often solve some spatial-temporal graph structures with hand-crafted energy terms, which are usually associated with features including appearance, boundary, motion and optical flows. Recently, deep-learning based methods were proposed due to its great success in image segmentation tasks~\cite{shelhamer2017fcn,Chen2016Deeplab,xu2016deep,xu2017deep}. Most of these methods~\cite{Caelles2017osvos,Perazzi2017masktrack,Cheng2017segflow,Jain_2017_CVPR,Yang2018osmn,voigtlaender2017online} build their model based on an image segmentation network and do not involve sequential modeling. Online learning~\cite{Caelles2017osvos} is commonly used to improve their performance. To make the model temporally consistent, the predicted mask of the previous frame is used as a guidance in~\cite{Perazzi2017masktrack,Yang2018osmn,Hu2017Maskrnn}. Other methods have been proposed to leverage spatial-temporal information. Jampani~\etal~\cite{Jampani2017vpn} use spatial-temporal consistency to propagate object masks over time. Tokmakov~\etal~\cite{Tokmakov2017memory} use a two-stream network to model objects' appearance and motion and use a recurrent layer to capture the evolution. However, due to the lack of training videos, they use a pretrained motion segmentation model~\cite{Tokmakov2017mpnet} and optical-flow model~\cite{Ilg2017flownet}, which leads to suboptimal results since the model is not trained end-to-end to best capture spatial-temporal features.

\section{\dataset}\label{sec:dataset}

To create our dataset, we first carefully select a set of object categories including animals (\eg \textit{ant, eagle, goldfish, person}), vehicles (\eg \textit{airplane, bicycle, boat, sedan}), accessories (\eg \textit{eyeglass, hat, bag}), common objects (\eg \textit{potted plant, knife, sign, umbrella}), and humans in various activities (\eg \textit{tennis, skateboarding, motorcycling, surfing}). The videos containing human activities have diversified appearance and motion, so instead of treating human videos as one class, we divide different activities into different categories. Most of these videos contain interactions between a person and a corresponding object, such as tennis racket, skateboard, motorcycle, etc. The entire category set includes 78 categories that covers diverse objects and motions, and should be representative for everyday scenarios. 




We then collect many high-resolution videos with the selected category labels from the large-scale video classification dataset YouTube-8M~\cite{abu2016youtube}. This dataset consists of millions of YouTube videos associated with more than 4,700 visual entities. We utilize its category annotations to retrieve candidate videos that we are interested in. Specifically, up to $100$ videos are retrieved for each category in our segmentation category set. There are several advantages to using YouTube videos to create our segmentation dataset. First, YouTube videos have very diverse object appearances and motions. Challenging cases for video object segmentation, such as occlusions, fast object motions and change of appearances, commonly exist in YouTube videos. Second, YouTube videos are taken by both professionals and amateurs and thus different levels of camera motions are shown in the crawled videos. Algorithms trained on such data could potentially handle camera motion better and thus are more practical. Last but not the least, many YouTube videos are taken by today's smart phone devices and there are demanding needs to segment objects in those videos for applications such as video editing and augmented reality. 

\begingroup
\setlength{\tabcolsep}{1pt}
\renewcommand{\arraystretch}{1}
\begin{figure*}[t]
\centering
 \begin{tabular}{@{}cccc@{}}
\includegraphics[width=.21\textwidth]{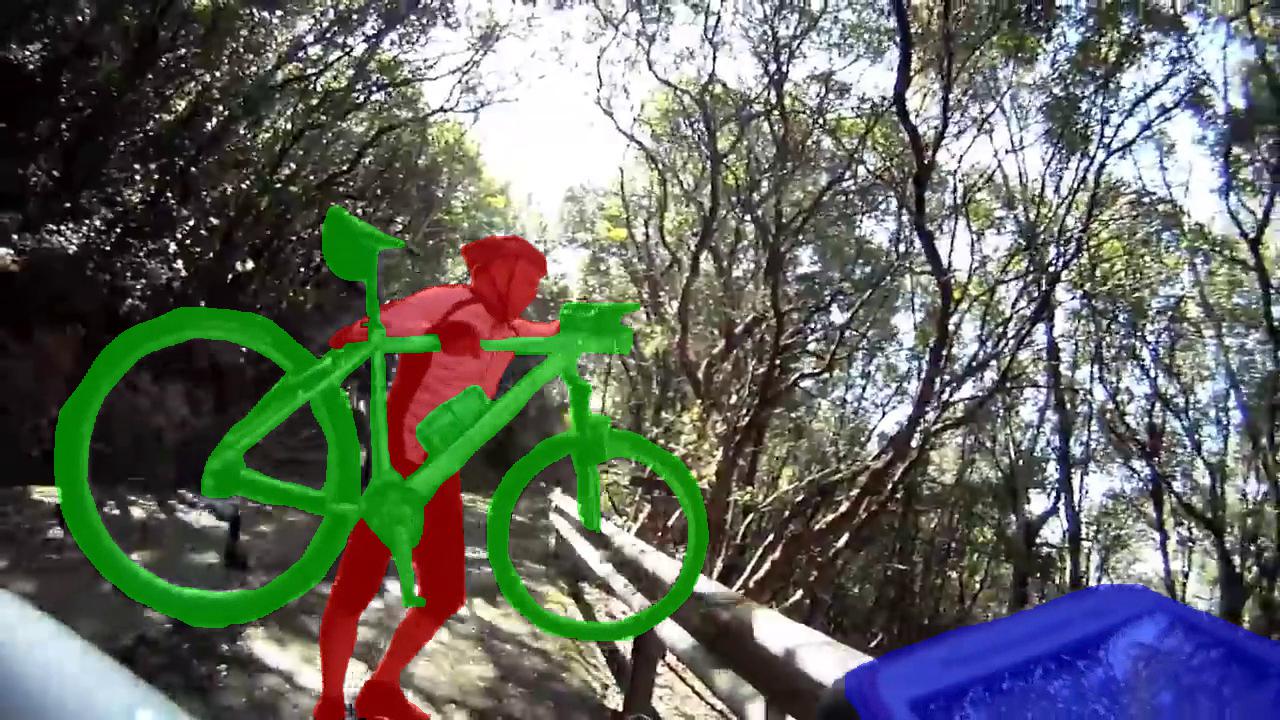}  &
\includegraphics[width=.21\textwidth]{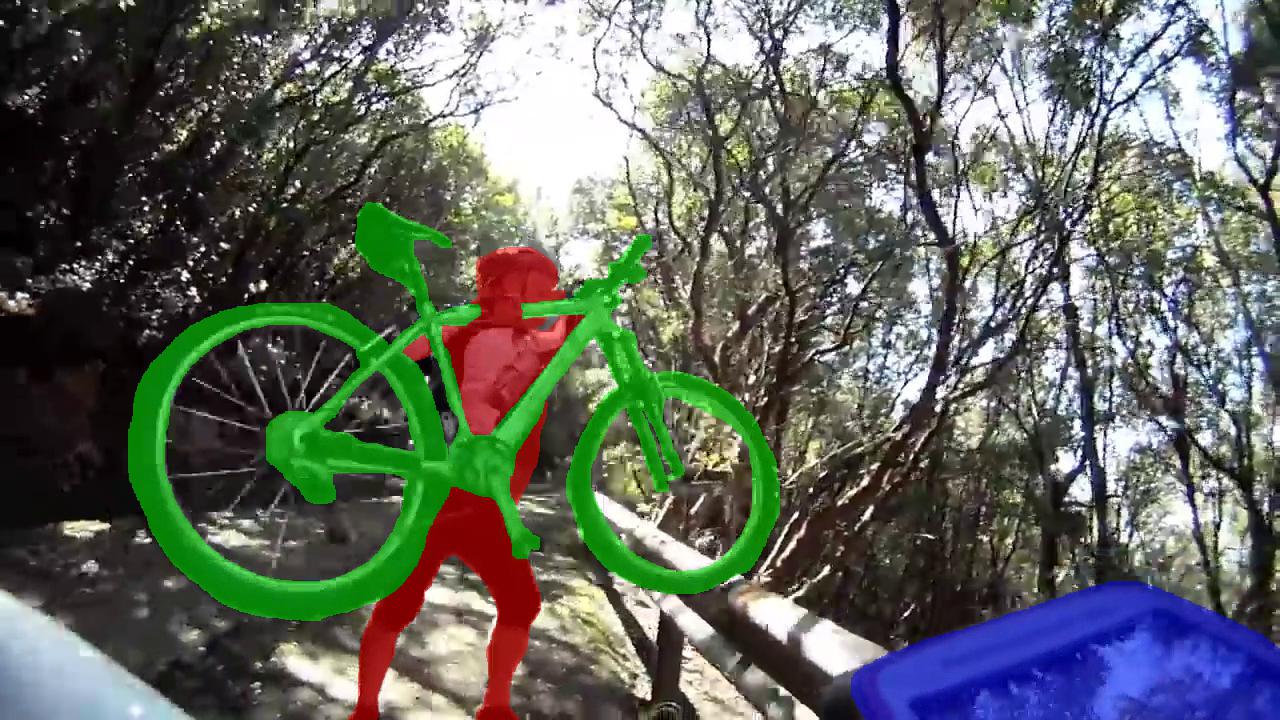}  &
 \includegraphics[width=.21\textwidth]{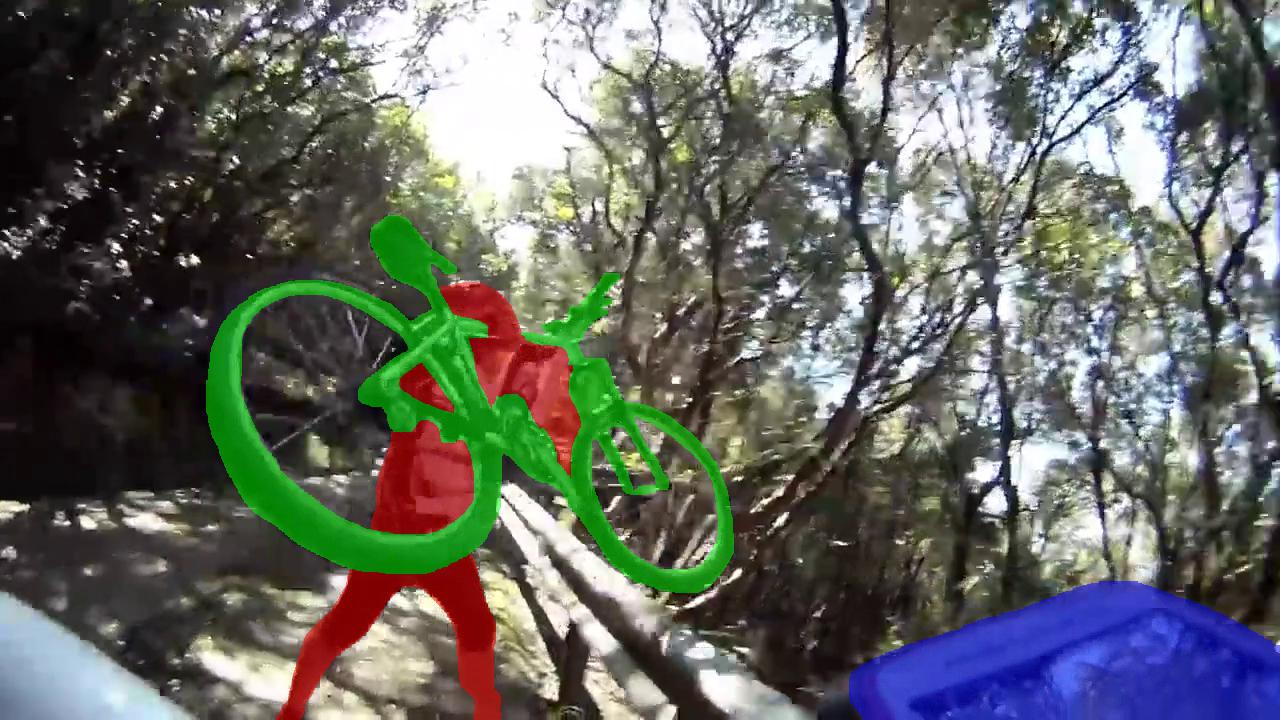}  &
 \includegraphics[width=.21\textwidth]{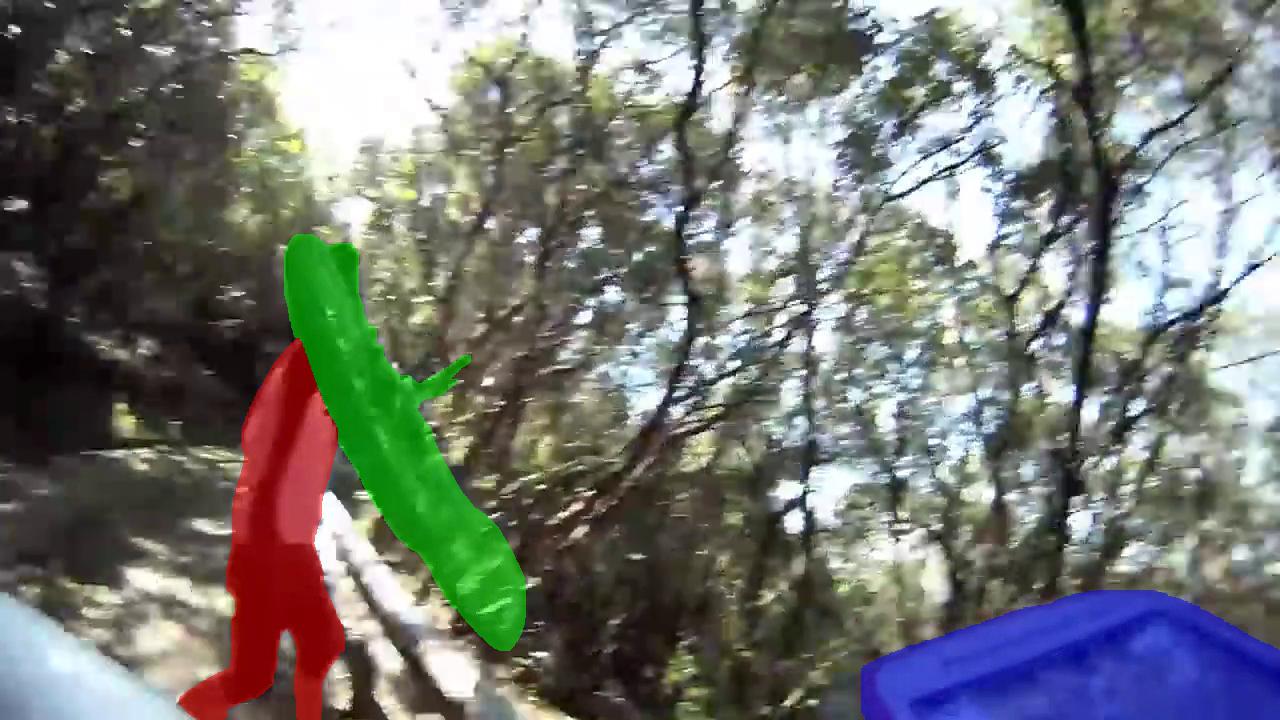}  \\
 \includegraphics[width=.21\textwidth]{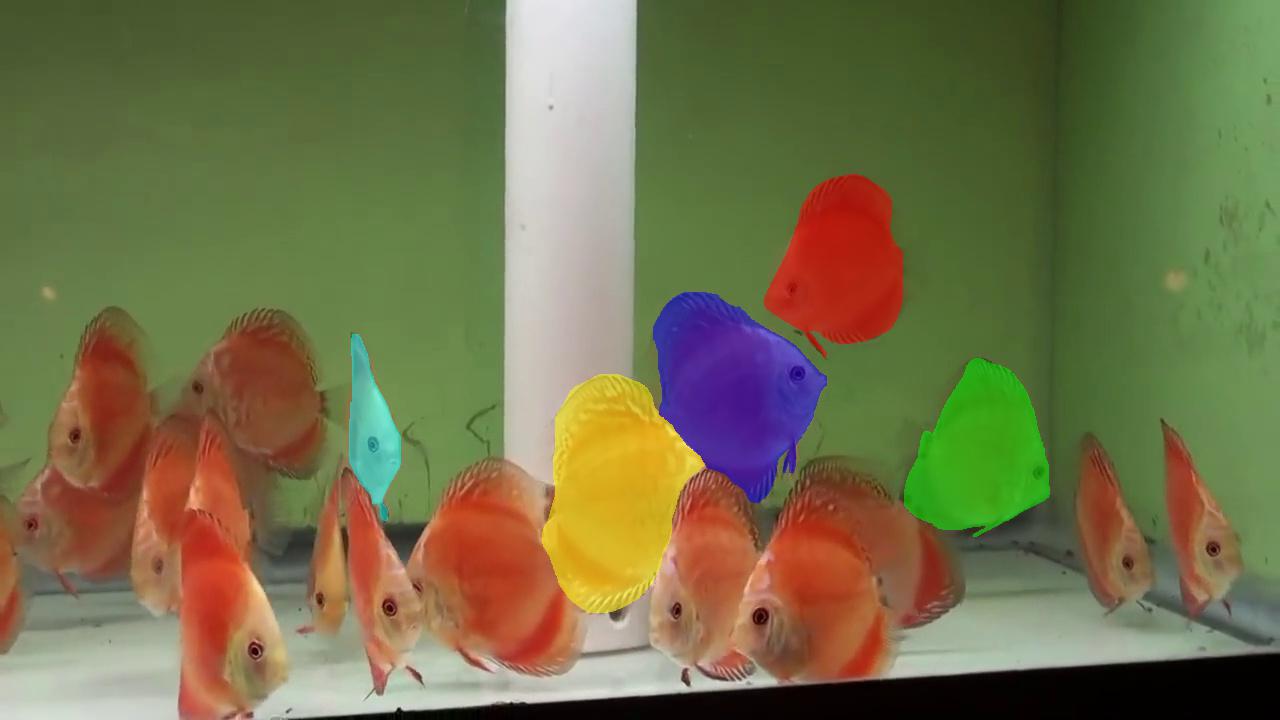}  &
\includegraphics[width=.21\textwidth]{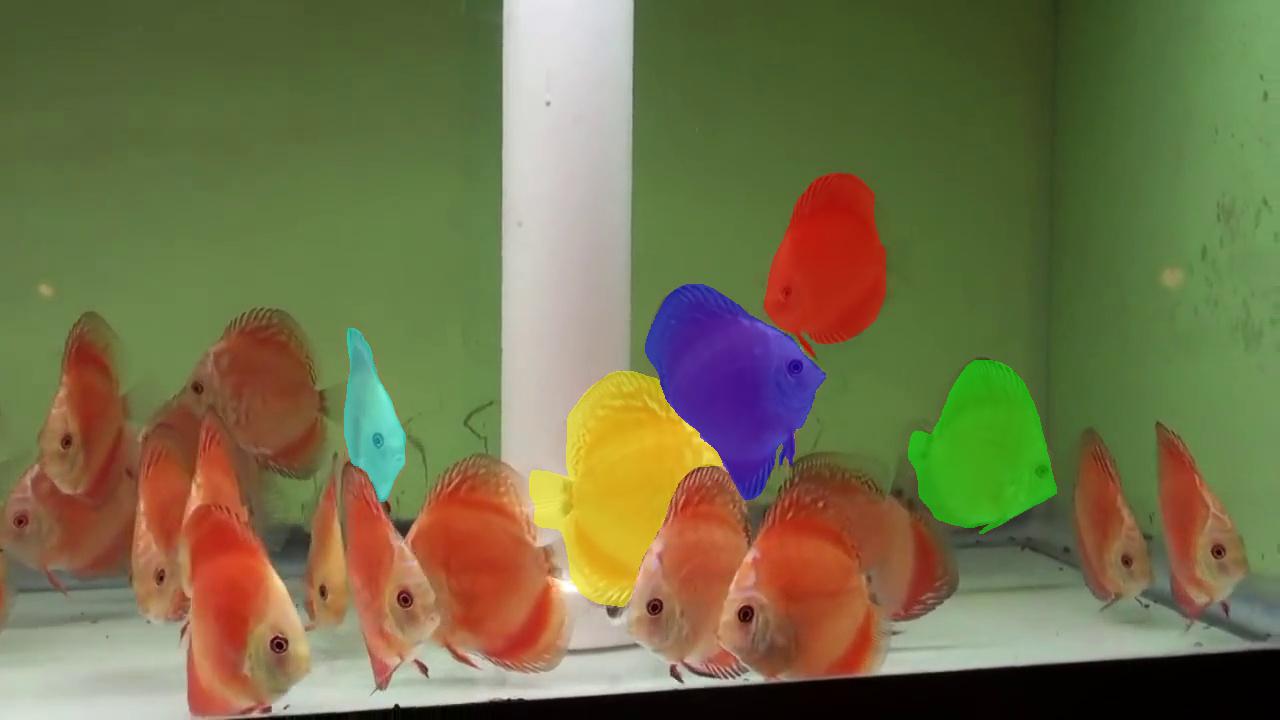}  &
 \includegraphics[width=.21\textwidth]{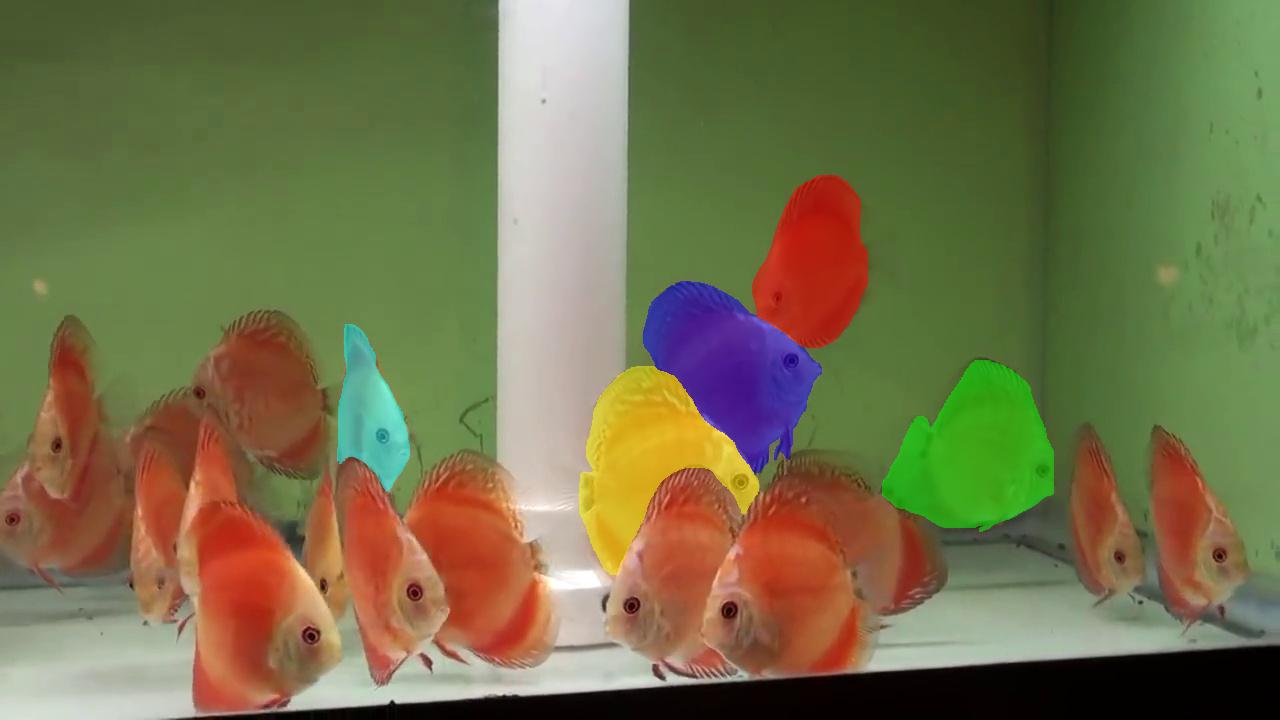}  &
 \includegraphics[width=.21\textwidth]{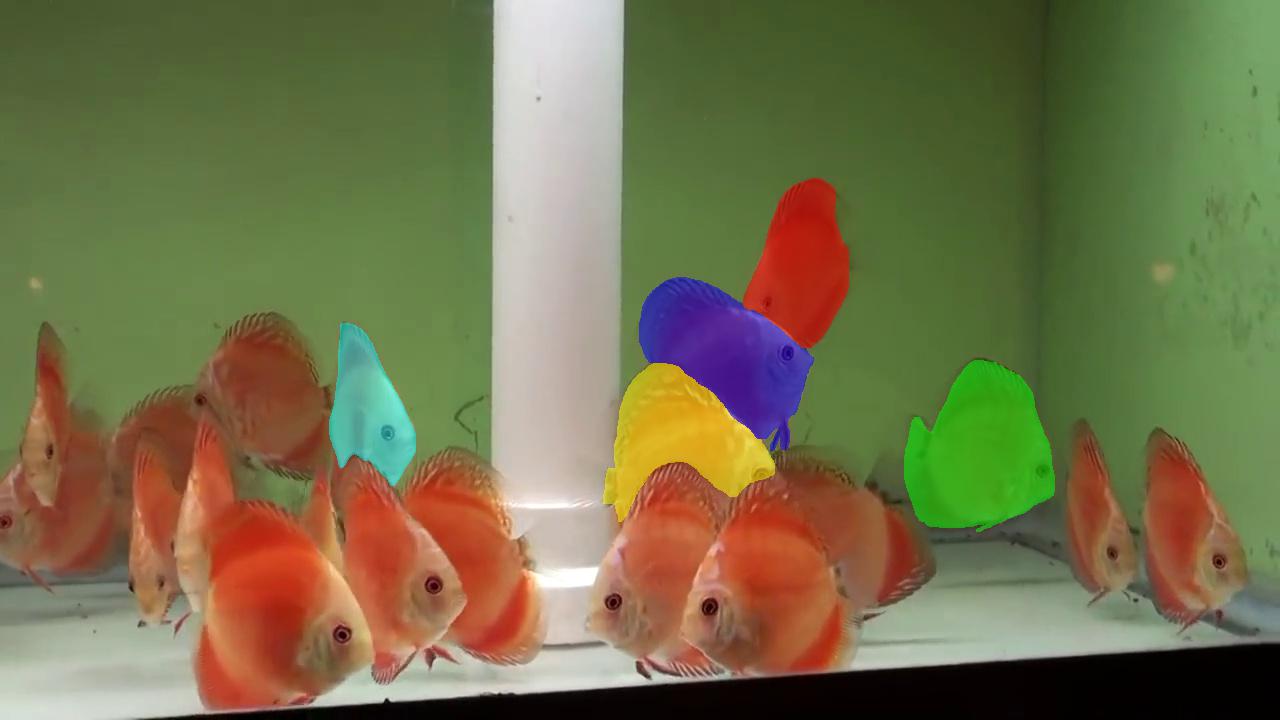}  
 \end{tabular}
 \caption{The ground truth annotations of sample video clips in our dataset. Different objects are highlighted with different colors.}
\label{fig:dataset_example}
\end{figure*}
\setlength{\tabcolsep}{1.4pt}
\endgroup

Since the retrieved videos are usually long (several minutes) and have shot transitions, we use an off-the-shelf video shot detection algorithm~\footnote{http://johmathe.name/shotdetect.html} to automatically partition each video into multiple video clips. We first remove the clips from the first and last 10\% of the video, since these clips have a high chance of containing introductory subtitles and credits lists. We then sample up to five clips with appropriate lengths (3$\sim$6 seconds) per video and manually verify that these clips contain the correct object categories and are useful for our task (\eg no scene transition, not too dark, shaky, or blurry).
After the video clips are collected, we ask human annotators to select up to five objects of proper sizes and categories per video clip and carefully annotate them (by tracing their boundaries instead of rough polygons) every five frames in a $30$fps frame rate, which results in a $6$fps sampling rate. Given a video and its category, annotators are first required to annotate objects belonging to that category. If the video contains other objects that belong to our 78 categories, we ask the annotators to label them as well, so that each video has multiple objects annotated. In human activity videos, both the human subject and the object he/she interacts with are labeled, \eg, both the person and the skateboard are required to be labeled in a ``skateboarding'' video. Some annotation examples are shown in Figure~\ref{fig:dataset_example}. Unlike dense per-frame annotation in previous datasets~\cite{jumpcut,Perazzi2016davis,Pont-Tuset2017davis}, we believe that the temporal correlation between five consecutive frames is sufficiently strong that annotations can be omitted for intermediate frames to reduce the annotation efforts. Such a skip-frame annotation strategy allows us to scale up the number of videos and objects under the same annotation budget, which are important factors for better performance. We find empirically that our dataset is effective in training different segmentation algorithm.


As a result, our dataset~\dataset~consists of 3,252 YouTube video clips and 133,886 object annotations, 33 and 10 times more than the best of the existing video object segmentation datasets, respectively (See Table~\ref{tab:dataset-cmp}). ~\dataset~is the largest dataset for video object segmentation to date. 

\section{Sequence-to-Sequence Video Object Segmentation}\label{sec:algorithm}

Based on our new dataset, we propose a new sequence-to-sequence video object segmentation algorithm. Different from existing approaches, our algorithm learns long-term spatial-temporal features directly from training data in an end-to-end manner, and the offline trained model is capable of propagating an initial object segmentation mask accurately by memorizing and updating the object charactersitics, including appearance, location and scale, and temporal movements, automatically over the entire video sequence.

\subsection{Problem formulation}

Let us denote a video sequence with $T$ frames as $\{\mathbf{x}_t |t\in[0,T-1]\}$ where $\mathbf{x}_t\in\mathbb{R}^{H\times W\times 3}$ is the RGB frame at time step $t$, and denote an initial binary object mask at time step 0 as $\mathbf{y}_0 \in\mathbb{R}^{H\times W}$. The target of video object segmentation is to predict the object mask automatically for the remaining frames from time step 1 to $T-1$, \ie \{$\mathbf{\hat y}_t|t\in[1,T-1]\}$. 

To obtain a predicted mask $\mathbf{\hat y}_t$ for $\mathbf{x}_t$, many existing deep learning methods only leverage information at time step 0 (\eg online learning or one-shot learning~\cite{Caelles2017osvos}) or time step $t-1$ (\eg optical flow~\cite{Perazzi2017masktrack}) while the long-term history information is totally dismissed. Their frameworks can be formulated as $\mathbf{\hat y}_t=\argmax_{\forall \mathbf{\bar y}_t}\mathbb{P}(\mathbf{\bar y}_t|\mathbf{x}_0,\mathbf{y}_0,\mathbf{x}_t)$ or $\mathbf{\hat y}_t=\argmax_{\forall \mathbf{\bar y}_t}\mathbb{P}(\mathbf{\bar y}_t|\mathbf{x}_0,\mathbf{y}_0,\mathbf{x}_t,\mathbf{x}_{t-1})$. They are effective when the object appearance is similar between time $0$ and time $t$ or when the object motion from time $t-1$ to $t$ can be accurately measured. However, these assumptions will be violated when the object has drastic appearance variation and rapid motion, which is often case in many real-world videos. In such cases, the history information of the object in all previous frames becomes critical and should be leveraged in an effective way. Therefore, we propose to solve a different objective function, \ie $\mathbf{\hat y}_t=\argmax_{\forall \mathbf{\bar y}_t}\mathbb{P}(\mathbf{\bar y}_t|\mathbf{x}_0,\mathbf{x}_1,...,\mathbf{x}_{t},\mathbf{y}_0)$, which can be transformed into a sequence-to-sequence learning problem.


\subsection{Our Algorithm}

Recurrent Neural Networks (RNN) has been adopted by many sequence-to-sequence learning problems because it is capable to learn long-term dependency from sequential data. LSTM~\cite{hochreiter1997long} as a special RNN structure solves vanishing or exploding gradients issue~\cite{bengio1994learning}. 
 A convolutional variant of LSTM (convolutional LSTM)~\cite{xingjian2015convolutional} is later proposed to preserve the spatial information of the data in the hidden states of the model. 


\begin{figure}[t]
    \centering
       \includegraphics[width=0.90\textwidth]{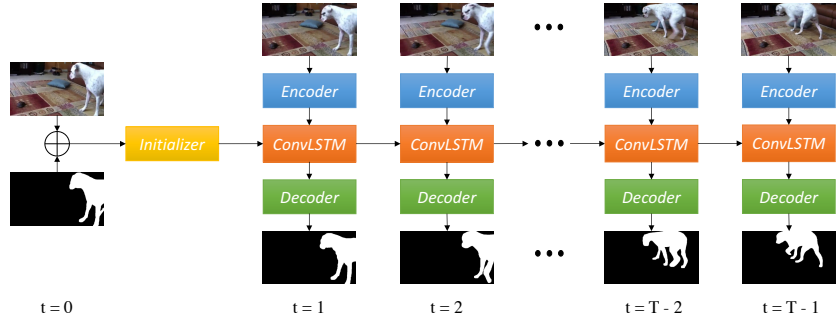}
    \caption{The framework of our algorithm. The initial information at time 0 is encoded by \textit{Initializer} to initialize \textit{ConvLSTM}. The new frame at each time step is processed by \textit{Encoder} and the segmentation result is decoded by \textit{Decoder}. \textit{ConvLSTM} is automatically updated over the entire video sequence.}
    \label{fig:framework}
\end{figure}

Our algorithm is inspired by the convolutional encoder-decoder LSTM structure~\cite{cho2014learning,sutskever2014sequence} which has achieved much success in machine translation,   
where an input sentence in language A is first encoded by a encoder LSTM and its outputs are fed into a decoder LSTM which can generate the desired output sentence in language B. 
In video object segmentation, it is essential to capture the object characteristics over time. To generate the initial states for our convolutional LSTM (\textit{ConvLSTM}), we use a feed-forward neural network to encode both the first image frame and the segmentation mask. Specifically, we concatenate the initial frame $\mathbf{x}_0$ and segmentation mask $\mathbf{y}_0$ and feed it into a trainable network, denoted as \textit{Initializer}, which outputs the initial memory state $\mathbf{c}_0$ and hidden state $\mathbf{h}_0$. These initial states capture object appearance, object location and scale. And they are are feed into \textit{ConvLSTM} for sequence learning. 

At time step $t$, frame $\mathbf{x}_t$ is first processed by a convolutional encoder, denoted as \textit{Encoder}, to extract feature maps $\mathbf{\tilde x}_t$. Then $\mathbf{\tilde x}_t$ is sent as the inputs of \textit{ConvLSTM}. The internal states $\mathbf{c}_t$ and $\mathbf{h}_t$ are automatically updated given the new observation $\mathbf{\tilde x}_t$, which capture the new characteristics of the object. The output $\mathbf{h}_t$ is passed into a convolutional decoder, denoted as \textit{Decoder}, to get the full-resolution segmentation results $\mathbf{\hat y}_t$. Binary cross-entropy loss is computed between $\mathbf{\hat y}_t$ and $\mathbf{y}_t$ during training process. The entire model is trained end-to-end using back-propagation to learn parameters for the \textit{Initializer} network, the \textit{Encoder} and \textit{Decoder} networks, and \textit{ConvLSTM} network. Figure~\ref{fig:framework} illustrates our sequence learning algorithm for video object segmentation. The learning process can be formulated as follows:  
\begin{align}
\mathbf{c}_0, \mathbf{h}_0 &= Initializer(\mathbf{x}_0,\mathbf{y}_0) \\
\mathbf{\tilde x}_t &= Encoder(\mathbf{x}_t) \\
\mathbf{c}_t, \mathbf{ h}_t  &= ConvLSTM(\mathbf{\tilde x}_t, \mathbf{c}_{t-1}, \mathbf{h}_{t-1}),\\
\mathbf{\hat y}_t &= Decoder(\mathbf{h}_t) \\
\mathcal{L} &= -(\mathbf{y_t}\log(\mathbf{\hat y_t})) + ((1-\mathbf{y_t})\log(1-\mathbf{\hat y_t})) 
\end{align}


\subsection{Implementation Details} \label{sec:implementation}

\paragraph{\textbf{Model structures}} Both our \textit{Initializer} and \textit{Encoder} use VGG-16~\cite{simonyan2014very} network structures. In particular, all the convolution layers and the first fully connected layer of VGG-16 are used as backbone for the two networks. The fully connected layer is transformed to a $1\times1$ convolution layer to make our model fully convolutional. On top of it, \textit{Initializer} has two additional convolution layers with ReLU~\cite{nair2010rectified} activation to produce $\mathbf{c_0}$ and $\mathbf{h_0}$ respectively. Each convolution layer has  $512$ $1\times 1$ filters.  The \textit{Encoder} has one additional convolution layer with ReLU activation which has $512$ $1\times1$ filters. The VGG-16 layers of the \textit{Initializer} and \textit{Encoder} are initialized with pre-trained VGG-16 parameters while the other layers are randomly initialized by Xavier~\cite{glorot2010understanding}. 

All the convolution operations of the \textit{ConvLSTM} layer use $512$ $3\times3$ filters, initialized by Xavier. Sigmoid activation is used for gate outputs and ReLU is used for state outputs (empirically we find ReLU activation produces better results than tanh activation for our model). Following~\cite{jozefowicz2015empirical}, we set the bias of the forget gate to be $1$s at initialization. 

The \textit{Decoder} has five upsampling layers with $5\times5$ kernel size and $512$, $256$, $128$, $64$ and $64$ filters respectively. The last layer of the \textit{Decoder} produces segmentation results, which has one $5\times5$ filter with sigmoid activation. All the parameters are initialized by Xavier.

\paragraph{\textbf{Training}} 
Our algorithm is trained on the~\dataset~training set. At each training iteration, our algorithm first randomly samples an object and $T$ $(5\sim 11)$ frames from a random training video sequence. Then the original RGB frames and annotations are resized to 256\(\times\)448 for memory and speed concern. At the early stage of training, we only select frames with ground truth annotation as our training samples so that the training loss can be computed and back-propagated at each time step. When the training losses become stable, we added frames without annotations to training data. For those frames without ground truth annotations, loss is set to be $0$. Adam~\cite{kingma2014adam} is used to train our network and the initial learning rate is set to $10^{-5}$, and our model converges in 80 epochs. 


\paragraph{\textbf{Inference}} Our offline-trained model is able to learn features for general object characteristics effectively. It is able to produce good segmentation results by directly applying it to a new test video with unseen categories. This is in contrast to recent state-of-the-art approaches, which have to fine-tune their models on each new test video over hundreds of iterations. In our experiments, we show that our algorithm without online learning can achieve comparable or better results compared to previous state of the arts with online learning, which implies much faster inference speed for practical applications. Neverthless, we find that the performance of our model can be further improved with online learning.  

\paragraph{\textbf{Online Learning}} Given a test video, we generate random pairs of online training examples $\{(\mathbf{x}^0, \mathbf{y}^0),(\mathbf{x}^1, \mathbf{y}^1)\}$ through affine transformations from $(\mathbf{x}_0, \mathbf{y}_0)$. We treat $(\mathbf{x}^0, \mathbf{y}^0)$ as the initial frame and mask and $(\mathbf{x}^1, \mathbf{y}^1)$ as the first frame and ground truth mask. We then fine tune our \textit{Initializer}, \textit{Encoder} and \textit{Decoder} networks on such randomly generated pairs. The parameters of \textit{ConvLSTM} are fixed as it models long-term spatial-temporal dependency that should be independent of object categories. 


\section{Experiments} \label{sec:exp}

In this section, we first evaluate our algorithm and recent state-of-the-art algorithms on our~\dataset~dataset. Then we compare our results on the DAVIS 2016 validation dataset~\cite{Perazzi2016davis}, which is an existing benchmark dataset for video object segmentation. Finally, we do an ablation study to explore the effect of data scale and model variants to our method.

\subsection{Experiment Settings}

We split the~\dataset~dataset of 3,252 videos into training (2,796), validation (134) and test (322) sets. To evaluate the generalization ability of existing approaches on unseen categories, the test set is further split into test-seen and test-unseen subsets. We first select 10 categories (\ie \textit{ant, bull riding, butterfly, chameleon, flag, jellyfish, kangaroo, penguin, slopestyle, snail}) as unseen categories during training and treat their videos as test-unseen set. The validation and test-seen subsets are created by sampling two and four videos per category, respectively. The rest of videos are the training set. We use the region similarity \(\mathcal{J}\) and the contour accuracy \(\mathcal{F}\) as the evaluation metrics as in~\cite{Perazzi2016davis}.

\subsection{\dataset}\label{sec:exp_new_data}

For fair comparison, we re-train previous methods (\ie SegFlow~\cite{Cheng2017segflow}, OSMN~\cite{Yang2018osmn}, MaskTrack~\cite{Perazzi2017masktrack}, OSVOS\cite{Caelles2017osvos} and OnAVOS~\cite{voigtlaender2017online}) on our training set with the same settings as our algorithm. One difference is that other methods leverage post-processing steps to achieve additional gains while our models do not.

\begin{figure}[t]
    \centering
    \begin{subfigure}[t]{0.5\textwidth}
        \centering
       \includegraphics[width=1\textwidth]{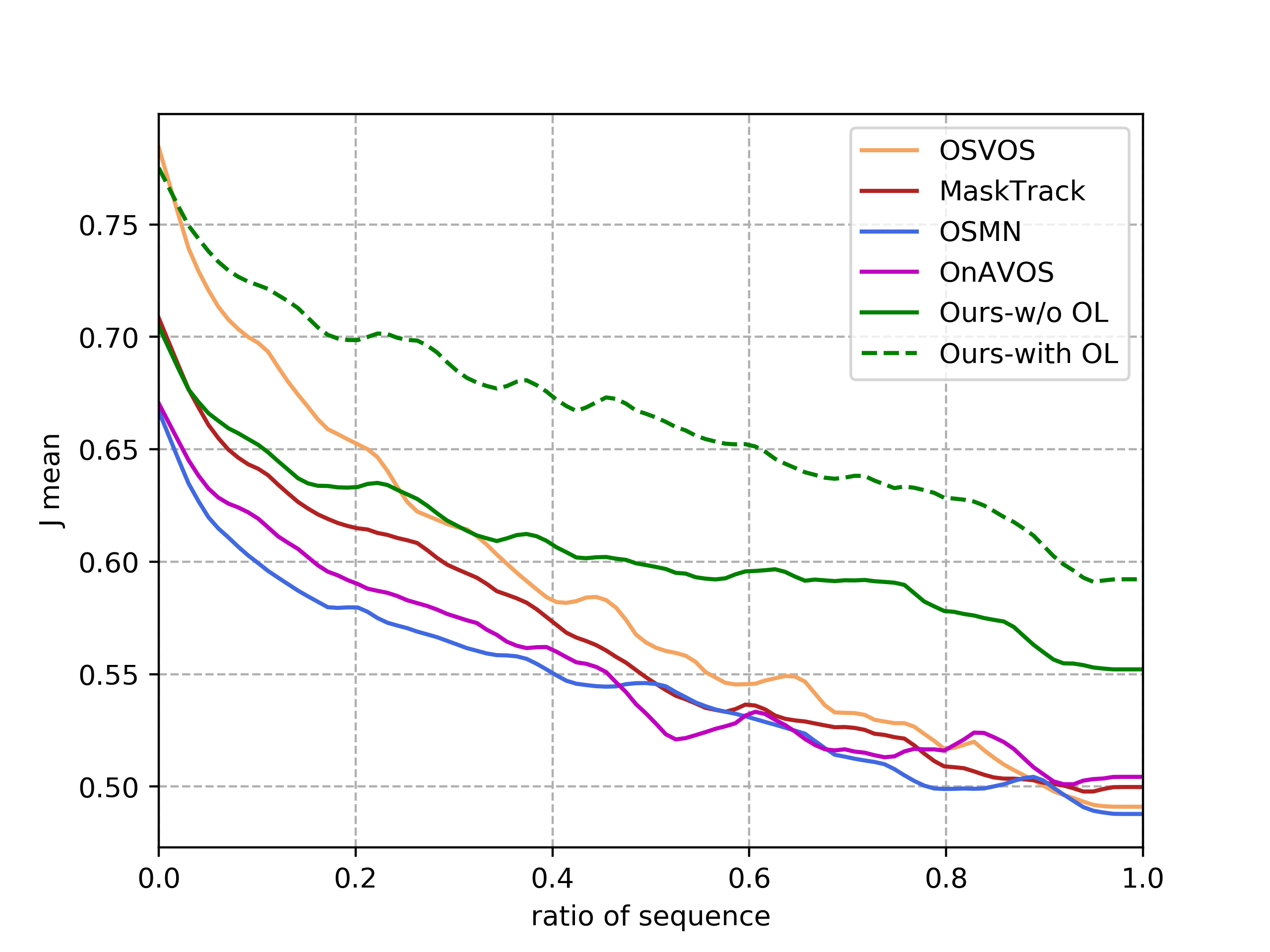}
        \caption{Seen categories}
    \end{subfigure}%
    ~ 
    \begin{subfigure}[t]{0.5\textwidth}
        \centering
        \includegraphics[width=1\textwidth]{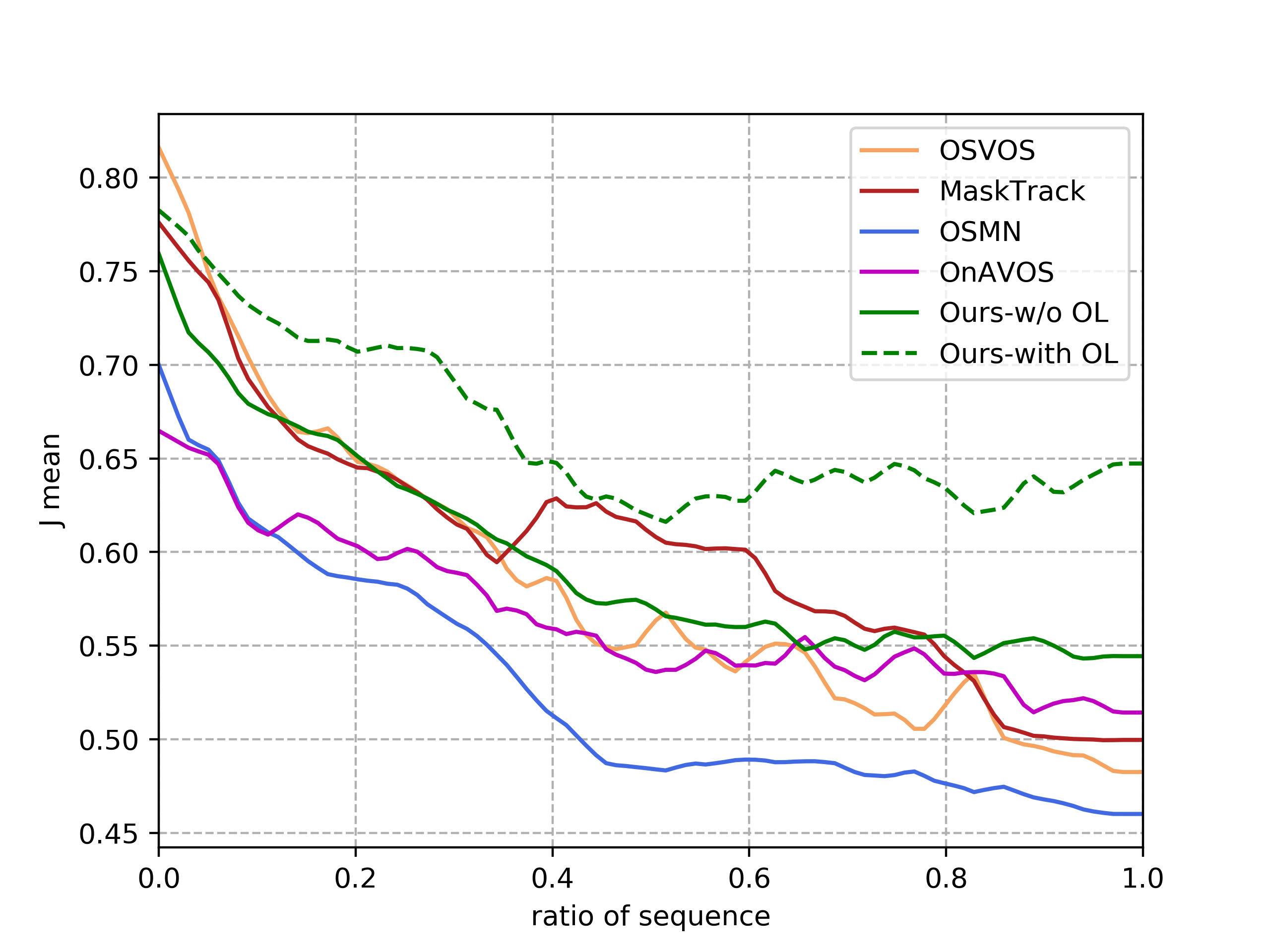}
        \caption{Unseen categories}
    \end{subfigure}
       
    \caption{The changes of $\mathcal{J}$ mean values over the length of video sequences.}
    \label{fig:j_over_time}
\end{figure}

The results are presented in Table~\ref{tab:res_2017}. All the comparison methods use static image segmentation models and four of them (\ie SegFlow, MaskTrack, OSVOS and OnAVOS) require online learning. Our algorithm leverages long-term spatial-temporal characteristics and achieves better performance even without online learinng (the second last row in Table~\ref{tab:res_2017}), which effectively demonstrates the importance of long-term spatial-temporal information for video object segmentation. With online learning, our model is further improved and achieves around $8\%$ absolute improvement over the best previous method OSVOS on $\mathcal{J}$ mean. Our method also outperforms previous methods on contour accuracy and decay rate with a large margin. Surprisingly, OnAVOS which is the best performing method on DAVIS does not achieve good results on our dataset. We believe the drastic appearance changes and complex motion patterns in our dataset makes the online adaptation fail in many cases.
Figure~\ref{fig:j_over_time} visualizes the changes of $\mathcal{J}$ mean over the duration of video sequences. Without online learning, our method is worse than online learning methods such as OSVOS at the first few frames since the object appearance usually has not changed too much from the initial frame and online learning is effective under such scenario. However, our method degrades slower than the other methods and starts to outperform OSVOS at around 25\% of the videos, which demonstrates that our method indeed propagates object segmentations more accurately over time than previous methods. With the help of online learning, our method outperforms previous methods in most parts of the video sequences, while maintaining a small decay rate.

\begin{table*}[t]
\centering
\caption{Comparisons of our approach and other methods on~\dataset~test set. The results in each cell show the test results for seen/unseen categories. ``OL'' denotes online learning. The best results are highlighted in bold.}
\label{tab:res_2017}
\begin{tabular}{|l|c|c|c|c|c|c|}
\hline
Method       & $\mathcal{J}$ mean$\uparrow$ & $\mathcal{J}$ recall$\uparrow$ & $\mathcal{J}$ decay$\downarrow$ & $\mathcal{F}$ mean$\uparrow$ & $\mathcal{F}$  recall$\uparrow$ & $\mathcal{F}$  decay$\downarrow$ \\ \hline
SegFlow~\cite{Cheng2017segflow}   & 40.4/38.5 & 45.4/41.7 & \textbf{7.2}/\textbf{8.4} & 35.0/32.7 & 35.3/32.1 & \textbf{6.9}/\textbf{9.1} \\ \hline
OSVOS~\cite{Caelles2017osvos}  & 59.1/58.8  & 66.2/64.5   & 17.9/19.5  & 63.7/63.9 & 69.0/67.9    &  20.6/23.0 \\ \hline
MaskTrack~\cite{Perazzi2017masktrack} &56.9/60.7     & 64.4/69.6   & 13.4/16.4     & 59.3/63.7   & 66.4/73.4   & 16.8/19.8    \\ \hline
OSMN~\cite{Yang2018osmn} & 54.9/52.9    & 59.7/57.6    & 10.2/14.6     & 57.3/55.2  & 60.8/58.0  & 10.4/13.8     \\ \hline
OnAVOS~\cite{voigtlaender2017online} & 55.7/56.8    & 61.6/61.5    & 10.3/{9.4}     & 61.3/62.3  & 66.0/67.3  & 13.1/12.8     \\ \hline
\textbf{Ours} (w/o OL)        &  60.9/60.1  &   70.3/71.2   & 7.9/12.9  & 64.2/62.3  &  73.0/71.4    &  9.3/14.5  \\ \hline
\textbf{Ours} (with OL)    &  \textbf{66.9}\textbf{/66.8}   &   \textbf{78.7}\textbf{/76.5}   &  10.2/9.5 & \textbf{74.1}\textbf{/72.3}  &   \textbf{82.8}\textbf{/80.5}   &  12.6/13.4  \\ \hline
\end{tabular}
\end{table*}

Next we compare the generalization ability of existing methods on unseen categories in Table~\ref{tab:res_2017}. Most methods have better performance on seen categories than unseen categories, which is expected. But the differences are not obvious, \eg usually within $2\%$ absolute differences on each metric. On one hand, it suggests that existing methods are able to alleviate the mismatch issue between training and test categories by approaches such as online learning. On the other hand, it also demonstrates the diverse training categories in~\dataset~helps different methods to generalize to new categories. Experiments on dataset scale in Section~\ref{sec:ablation} further suggests the power of data scale on our model. Compared to other single-frame based methods, OSMN has a more obvious degradation on unseen categories since it does not use online learning. Our method without online learning does not have the issue since it leverages spatial-temporal information which is more robust to unseen categories.    
MaskTrack and OnAVOS have better performance on unseen than seen categories. We believe that they benefit from the guidance of previous segmentation or online adaption, which have advantages to deal with videos with slow motion. There are indeed several objects with slow motion in the unseen categories such as \textit{snail} and \textit{chameleon}.

\begingroup
\setlength{\tabcolsep}{1pt}
\renewcommand{\arraystretch}{1}
\begin{figure*}[t]
\centering
 \begin{tabular}{@{}ccccc@{}}
\includegraphics[width=.19\textwidth]{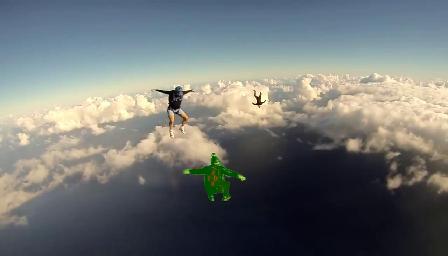}  &
\includegraphics[width=.19\textwidth]{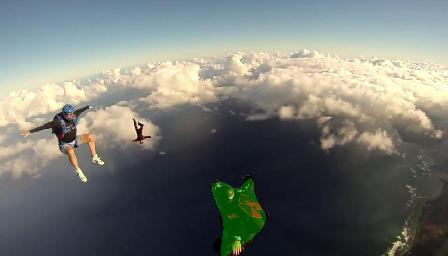}  &
\includegraphics[width=.19\textwidth]{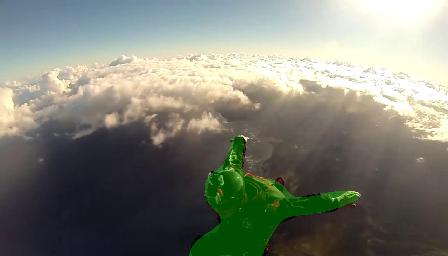}  &
 \includegraphics[width=.19\textwidth]{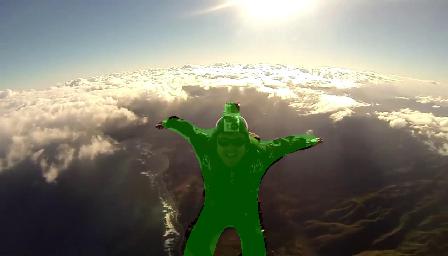}  &
 \includegraphics[width=.19\textwidth]{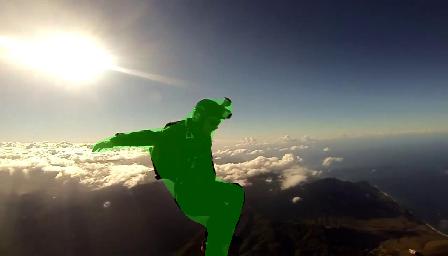}  \\
 \includegraphics[width=.19\textwidth]{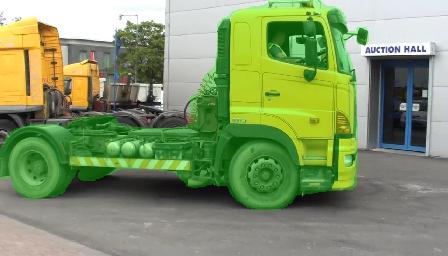}  &
\includegraphics[width=.19\textwidth]{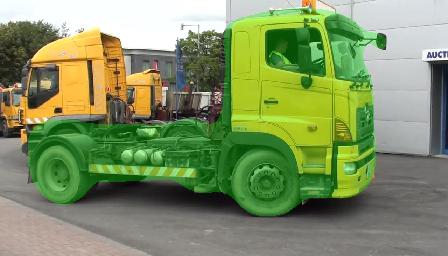}  &
\includegraphics[width=.19\textwidth]{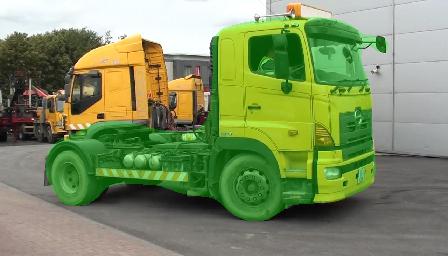}  &
 \includegraphics[width=.19\textwidth]{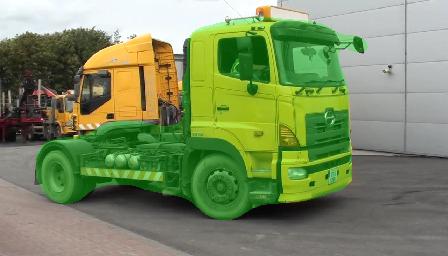}  &
 \includegraphics[width=.19\textwidth]{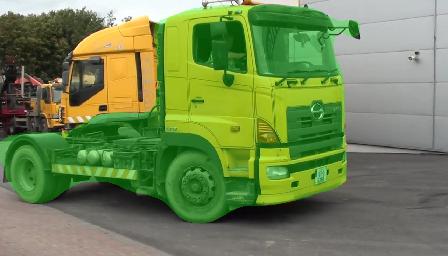}  \\
 \includegraphics[width=.19\textwidth]{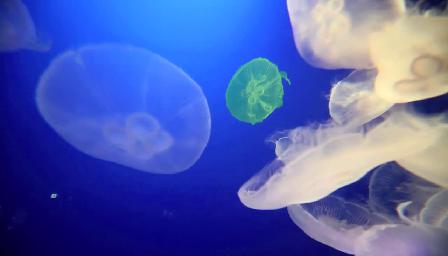}  &
\includegraphics[width=.19\textwidth]{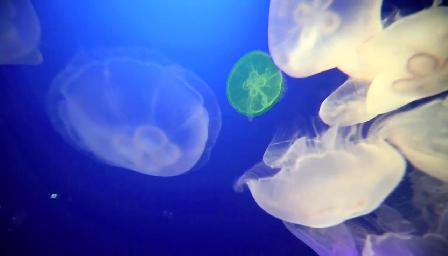}  &
\includegraphics[width=.19\textwidth]{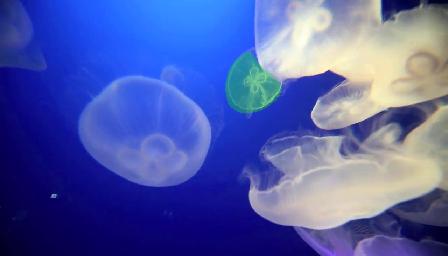}  &
 \includegraphics[width=.19\textwidth]{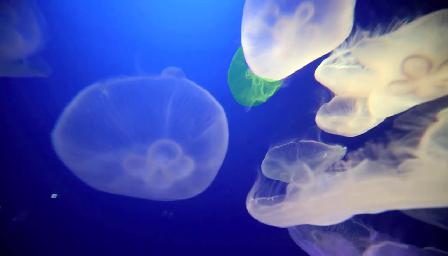}  &
 \includegraphics[width=.19\textwidth]{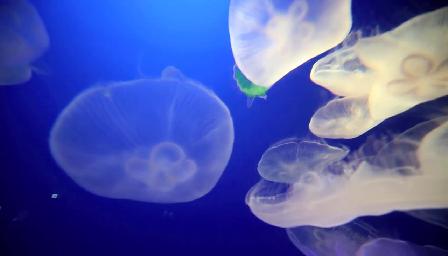}  \\
 \includegraphics[width=.19\textwidth]{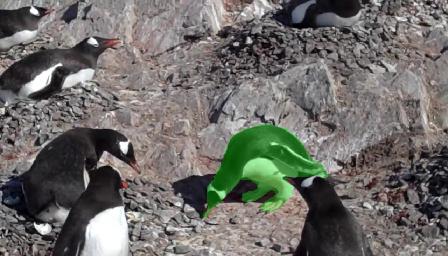}  &
\includegraphics[width=.19\textwidth]{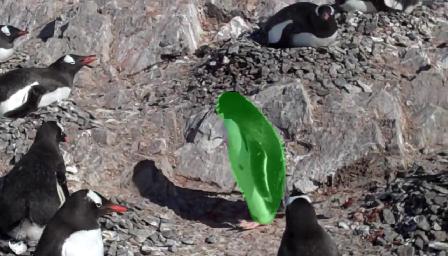}  &
\includegraphics[width=.19\textwidth]{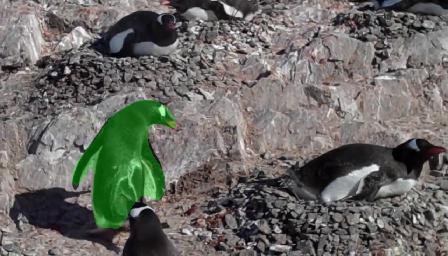}  &
 \includegraphics[width=.19\textwidth]{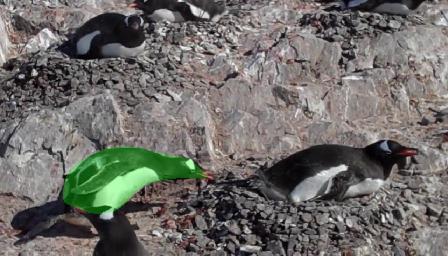}  &
 \includegraphics[width=.19\textwidth]{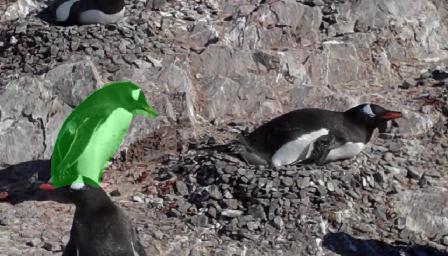}  
 \end{tabular}
 \caption{Some visual results produced by our model without online learning on the~\dataset~test set. The first column shows the initial ground truth object segmentation (green color) while the second to the last column are predictions.  }
\label{fig:youtube-vos-res}
\end{figure*}
\setlength{\tabcolsep}{1.4pt}
\endgroup

Some test results produced by our model without online learning are visualized in Figure~\ref{fig:youtube-vos-res}. The first two rows are from seen categories while the last two rows are from unseen categories. In addition, each example represents some challenging cases in video object segmentation. For example, the person in the first example has large changes in appearance and illumination. The second and third examples both have multiple similar objects and heavy occlusions. The last example has strong camera motion and the penguin changes its pose frequently. Our model obtains accurate results on all the examples, which demonstrates the effectiveness of spatial-temporal features learned from large-scale training data.

\subsection{DAVIS 2016}

DAVIS 2016 is a popular prior benchmark dataset for video object segmentation. To evaluate our algorithm, we first fine-tune our pretrained model in 200 epochs on the DAVIS training set which contains 30 videos. The comparison results between our fine-tuned models and previous methods are shown in Table~\ref{tab:res_davis_2016}.  

\begin{table*}[t]
\centering
\caption{Comparisons of our approach and previous methods on the DAVIS 2016 dataset. Different components used in each algorithm are marked. ``OL" denotes online learning. ``PP'' denotes post processing by CRF~\cite{Krahenbuhl2011crf} or Boundary Snapping~\cite{Caelles2017osvos}. ``OF'' denotes optical flows. ``RNN'' denotes RNN and its variants. }
\label{tab:res_davis_2016}
\begin{tabular}{|l|c|c|c|c|c|c|}
\hline
Method & OL & PP & OF & RNN & mean IoU($\%$) &  Speed(s) \\ \hline
BVS~\cite{Tsai2016objflow}&  -   &  \xmark & \xmark &  -    &   60.0 & 0.37  \\ \hline
OFL~\cite{Marki2016bilateral}  &  -   & \cmark  & \cmark &  -    &   68.0  & 42.2 \\ \hline
SegFlow~\cite{Cheng2017segflow}   &  \cmark   & \cmark   & \cmark &  \xmark    &  76.1  & 7.9  \\ \hline
MaskTrack~\cite{Perazzi2017masktrack} &  \cmark    &  \cmark  &  \xmark    &  \xmark  & 79.7  &  12   \\ \hline
OSVOS~\cite{Caelles2017osvos}  &  \cmark & \cmark  & \xmark  & \xmark  & 79.8 &  10    \\ \hline
OnAVOS~\cite{voigtlaender2017online} &  \cmark & \cmark  & \xmark  & \xmark  & \textbf{85.7} &  13    \\ \hline
OSMN~\cite{Yang2018osmn}       &  \xmark  &  \xmark    &  \xmark & \xmark &  74.0    &  \textbf{0.14}  \\ \hline
VPN~\cite{Jampani2017vpn} &  \xmark  &   \xmark   & \xmark  & \xmark  & 70.2  &   0.63   \\ \hline
ConvGRU~\cite{Tokmakov2017memory}   &  \xmark   & \cmark   & \cmark &  \cmark    &   75.9  & 20 \\ \hline
\textbf{Ours}       &  \xmark  &   \xmark   & \xmark  & \cmark  & 76.5 & 0.16 \\ \hline
\textbf{Ours}    &  \cmark  &  \xmark    & \xmark  & \cmark &   {79.1}   &  9  \\ \hline
\end{tabular}
\end{table*}

BVS and OFL are based on hand-crafted features and graphical models, while the rest are all deep learning based methods. Among the methods~\cite{Perazzi2017masktrack,Caelles2017osvos,voigtlaender2017online,Yang2018osmn} using image segmentation frameworks, OnAVOS achieves the best performance. However, its online adaption process makes the inference pretty slow ($\sim$13s per frame). Our model without online learning (the second last row) achieves comparable results to other online learning methods without post-processing (\eg MaskTrack 69.8\% and OSVOS 77.4\%), but with a significant speed-up (60 times faster). Previous methods using spatial-temporal information including SegFlow, VPN and ConvGRU get inferior results compared to ours. Among them, ConvGRU is most related to ours since it also incorporates RNN memory cells in its model. However, it is an unsupervised methods to only segment moving foreground, while our method can segment arbitrary objects given the mask supervision.
Finally, online learning helps our model segment object boundary more accurately. Figure~\ref{fig:davis-res} shows such an example. 

\begingroup
\setlength{\tabcolsep}{1pt}
\renewcommand{\arraystretch}{1}
\begin{figure*}[t]
\centering
 \begin{tabular}{@{}ccccc@{}}
 \includegraphics[width=.19\textwidth]{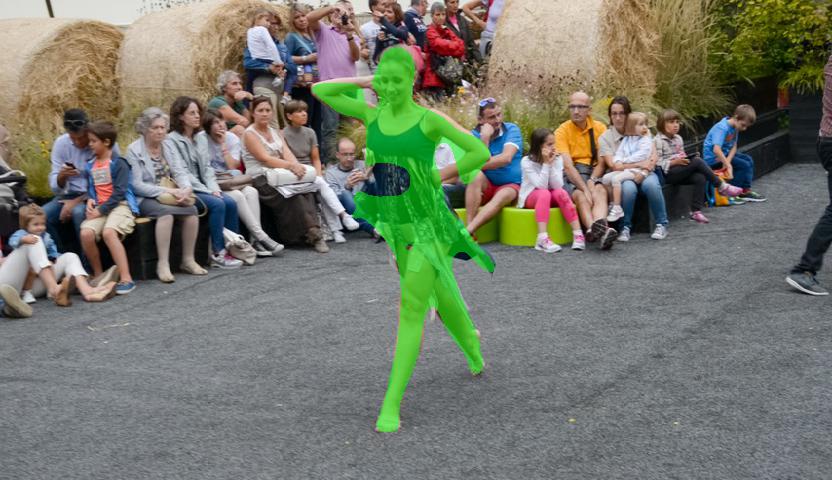}  &
\includegraphics[width=.19\textwidth]{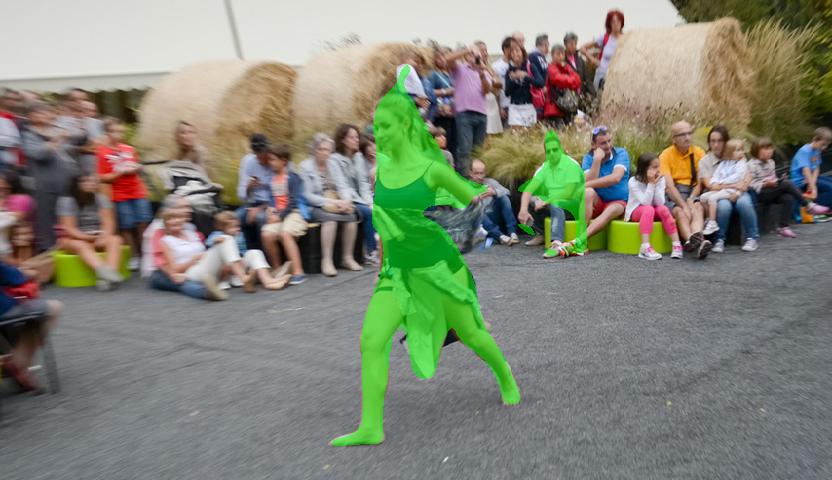}  &
\includegraphics[width=.19\textwidth]{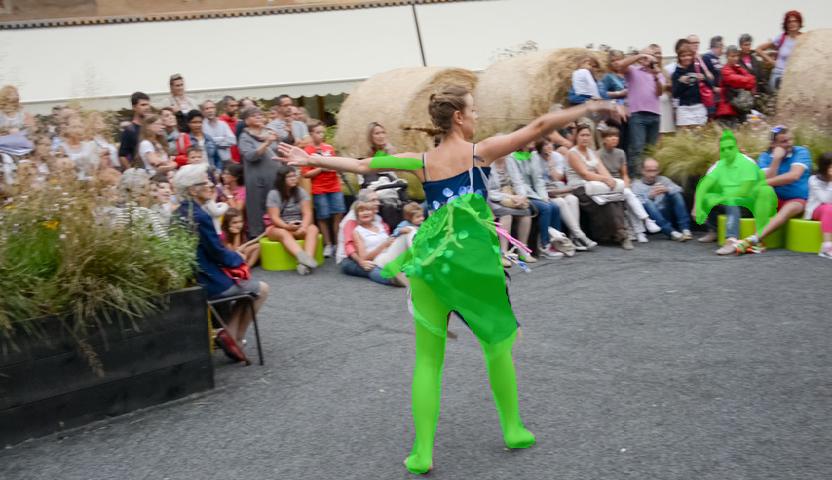}  &
 \includegraphics[width=.19\textwidth]{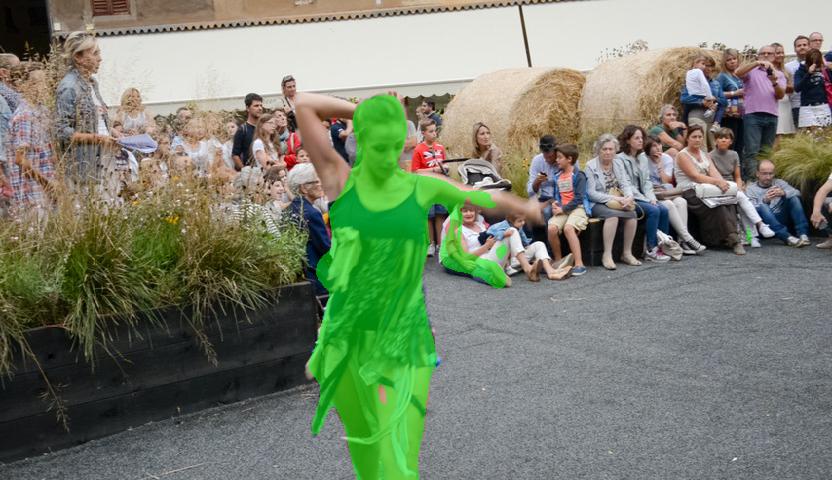}  &
 \includegraphics[width=.19\textwidth]{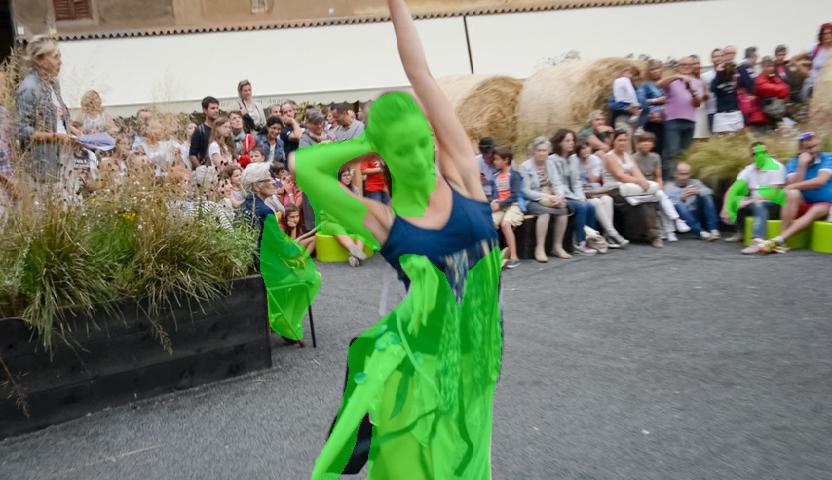}  \\
 \includegraphics[width=.19\textwidth]{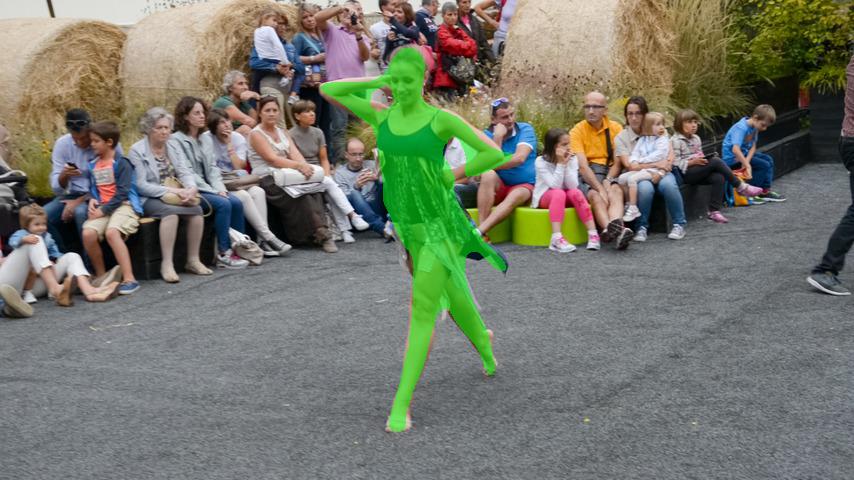}  &
\includegraphics[width=.19\textwidth]{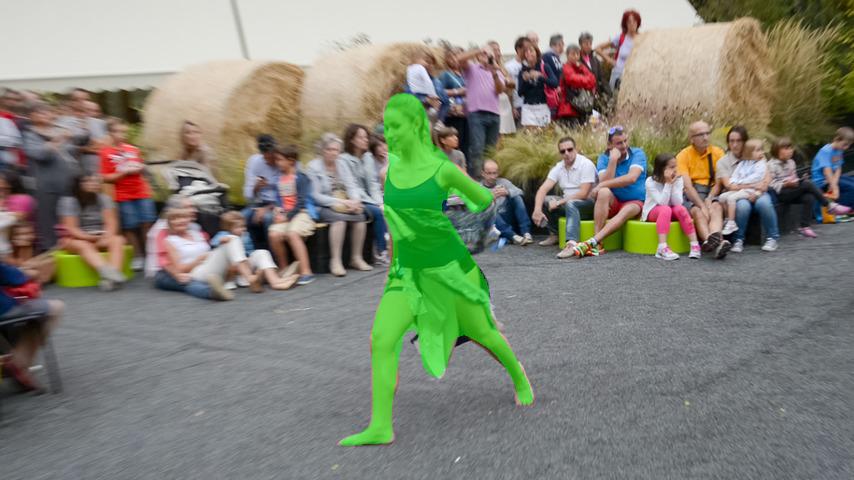}  &
\includegraphics[width=.19\textwidth]{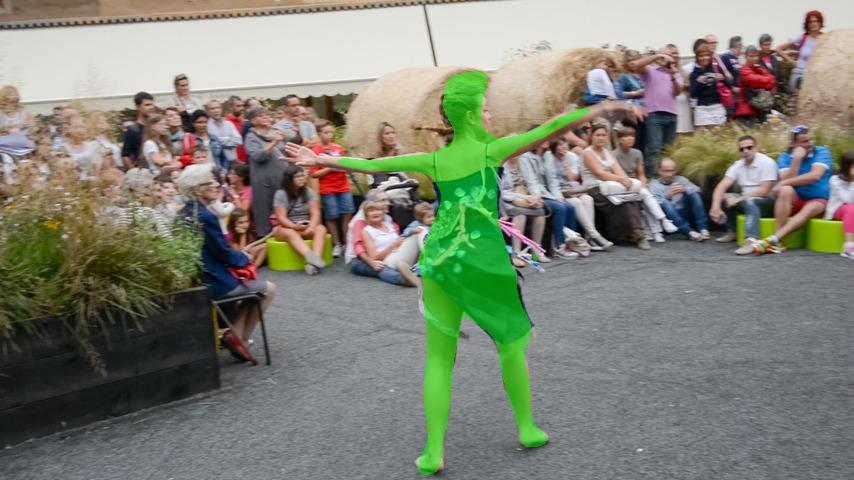}  &
 \includegraphics[width=.19\textwidth]{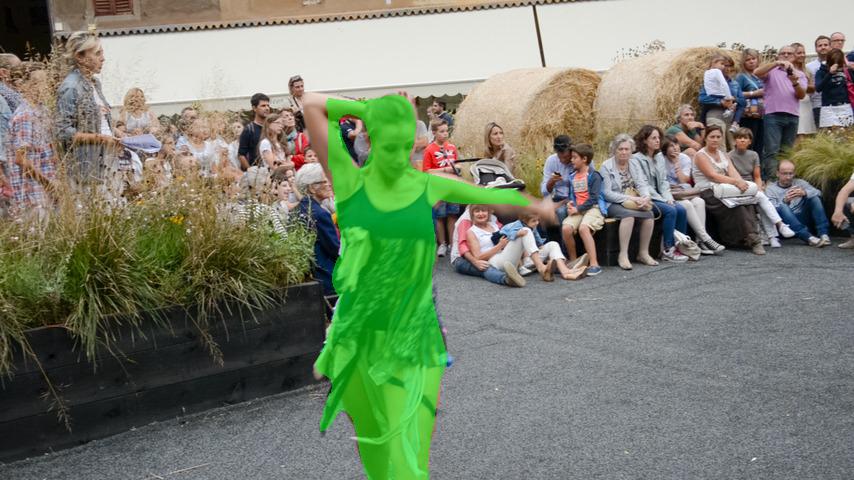}  &
 \includegraphics[width=.19\textwidth]{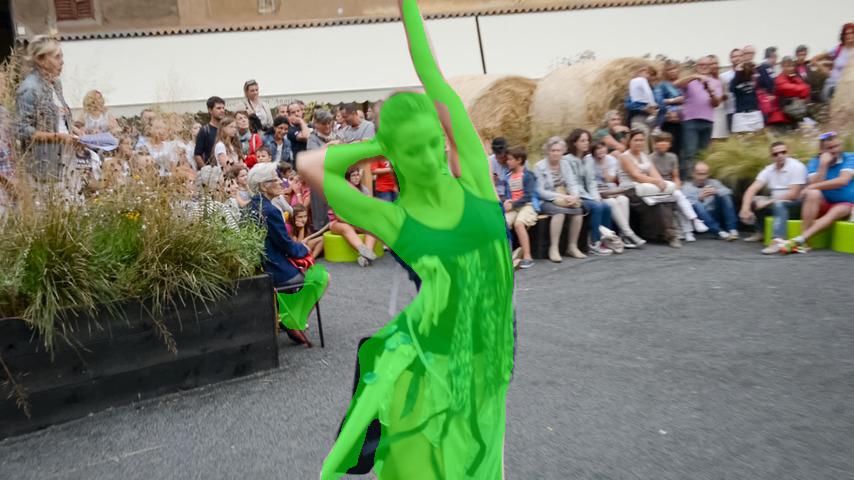}  
 \end{tabular}
 \caption{The comparison results between our model without online learning (upper row) and with online learning (bottom row). Each column shows predictions of the two models at the same frame. }
\label{fig:davis-res}
\end{figure*}
\setlength{\tabcolsep}{1.4pt}
\endgroup

To demonstrate the scale limitation of existing datasets, we train our models on three different settings and evaluate on DAVIS 2016. 
\begin{itemize}
\item Setting 1: We train our model from scratch on the 30 training videos. 
\item Setting 2: We train our model from scratch on the 30 training videos, plus all the videos from the SegTrackv2, JumpCut and YoutubeObjects datasets, which results in a total of 192 training videos.
\item Setting 3: Following the idea of ConvGRU, we use a pretrained object segmentation model DeepLab ~\cite{Chen2016Deeplab} as our \textit{Encoder} and train the other parts of our model on the 30 training videos.
\end{itemize}
Our models trained on setting 1 and 2 only get $51.3\%$ and $51.9\%$ mean IoU, which suggests that existing video object segmentation datasets do not have sufficient data to train our models. Therefore our \dataset~dataset is one of the key elements for the success of our algorithm. In addition, there is only little improvement by adding videos from the SegTrackv2, JumpCut and YoutubeObjects datasets, 
which suggests that the small scale is not the only problem for previous datasets. For example, videos in the three datasets usually only have one main foreground. SegTrackv2 has low-resolution videos. The annotation of YoutubeObjects videos is not accurate along object boundaries, \etc   However, our \dataset~dataset is carefully created to avoid all these problems. 
Setting 3 is a common detour for existing methods to bypass the data-insufficiency issue, \ie using pre-trained models on other large-scale datasets to reduce the parameters to be learned for their models. However, our model using this strategy gets even worse results ($45.6\%$) than training from scratch, which suggests that spatial-temporal features cannot be trivially transfered from representations learned from static images. Thus large scale training data such as our dataset is essential to learn spatial-temporal representation for video object segmentation.

\subsection{Ablation study} \label{sec:ablation}

In this subsection, we perform an ablation study on the~\dataset~dataset to evaluate different variants of our algorithm.

\begin{table*}[t]
\centering
\caption{The effect of data scale on our algorithm. We use different portions of training data to train our models and evaluate on the~\dataset~test set.}
\label{tab:ablation}
\begin{tabular}{|l|c|c|c|c|c|c|}
\hline
Scale      & $\mathcal{J}$ mean$\uparrow$ & $\mathcal{J}$ recall$\uparrow$ & $\mathcal{J}$ decay$\downarrow$ & $\mathcal{F}$ mean$\uparrow$ & $\mathcal{F}$  recall$\uparrow$ & $\mathcal{F}$  decay$\downarrow$ \\ \hline
25$\%$ &  46.7/40.1    &  53.5/45.6  & 8.3/13.6  &   46.7/40.0   &  52.2/41.6   &  8.5/13.2  \\ \hline
50$\%$ &  51.5/50.3   &  59.2/58.8    &  10.3/13.1  & 51.8/50.2  &  59.5/55.8  &  11.1/13.3    \\ \hline
75$\%$ &   56.8/56.0  &  65.7/67.1   &   7.6/10.0   & 59.6/56.3  & 68.8/64.1  &  8.5/11.1    \\ \hline
100$\%$  &  60.9/60.1  &   70.3/71.2   & 7.9/12.9  & 64.2/62.3  &  73.0/71.4    &  9.3/14.5  \\ \hline
\end{tabular}
\end{table*}

\paragraph{\textbf{Dataset scale.}}
Since the dataset scale is very important to our models, we train several models on different portions of the training set of~\dataset~to explore the effect of data. Specifically, we randomly select $25\%$, $50\%$ and $75\%$ of the training set and retrain our models from scratch. The results are listed in Table~\ref{tab:ablation}. It can be seen that using only $25\%$ of the training videos ($\sim$700 videos) drops the performance almost $30\%$ from the original model. In addition, the performance of the model on unseen categories are much worse than its performance on seen categories, which suggests that the model fails to capture general features for objectness. Since the scale of adding all the videos from all existing datasets is still much less than 700 videos, there is no doubt that existing datasets cannot satisfy the needs of our algorithm. With more and more training videos, our algorithm is improved rapidly, which well demonstrates the importance of large-scale data on our algorithm. We can see the trend of accuracies for $100\%$ data still have not reached a plateau. We are working on collecting more data to explore the impact of data on the algorithm further.

\paragraph{\textbf{\textit{Initializer} variants.}}
The \textit{Initializer} in our original model is a VGG-16 network which encodes a RGB frame and an object mask and outputs initial hidden states of \emph{ConvLSTM}. We would like to explore using the object mask directly as the initial hidden states of \emph{ConvLSTM}. We train an alternative model by removing the \textit{Initializer} and directly using the object mask as the hidden states, \ie the object mask is reshaped to match the size of the hidden states. The $\mathcal{J}$ mean of the adapted model are $45.1\%$ on the seen categories and $38.6\%$ on the unseen categories. This suggests that the object mask alone does not have enough information for localizing the object.

\paragraph{\textbf{\textit{Encoder} variants.}}
The \textit{Encoder} in our original model receives a RGB frame as input at each time step. Alternatively, we can use the segmentation mask of the previous step as additional inputs to explicitly provide extra information to the model, similar to MaskTrack~\cite{Perazzi2017masktrack}. In this way, our \textit{Initializer} and \textit{Encoder} can be replaced with a single VGG-16 network since the inputs at every time step have same dimensions and similar meaning. However, such a framework potentially has the error-drifting issue since segmentation mistakes made at previous steps will be propagated to the current step. 

In the early stage of training, the model is unable to predict good segmentation results. Therefore we use the ground truth annotation of the previous step as the input. Such strategy is known as teacher forcing~\cite{williams1989learning} which can make the training faster. After the training losses become stable, we replace the ground truth annotation with the model's prediction of the previous step so that the model is forced to correct its own mistakes. Such a strategy is known as curriculum learning~\cite{bengio2015scheduled}. Empirically we find that both the two strategies are important to make the model to work well. The $\mathcal{J}$ mean results of the model are $59.4\%$ on the seen categories and $60.7\%$ on the unseen categories, which is similar to our original model.

\section{Conclusion} \label{sec:conclusion}
In this work, we introduce the largest video object segmentation dataset (\dataset) to date. The new dataset, much larger than existing datasets in terms of number of videos and annotations, allows us to design a new deep learning algorithm to explicitly model long-term spatial-temporal dependency from videos for segmentation in an end-to-end learning framework. Thanks to the large scale dataset, our new algorithm achieves better or comparable results compared to existing state-of-the-art approaches. We believe the new dataset will foster research on video-based computer vision in general.

\section{Acknowledgement}
This research was partially supported by a gift funding from Snap Inc. and UIUC Andrew T. Yang Research and Entrepreneurship Award to Beckman Institute for Advanced Science \& Technology, UIUC. 

\bibliographystyle{splncs04}
\bibliography{egbib}
\end{document}